\documentclass{article}

\PassOptionsToPackage{numbers,sort}{natbib}
\usepackage[preprint]{neurips_2023}

\usepackage[utf8]{inputenc} 
\usepackage[T1]{fontenc}    
\usepackage{hyperref}       
\usepackage{url}            
\usepackage{booktabs}       
\usepackage{amsfonts}       
\usepackage{nicefrac}       
\usepackage{microtype}      
\usepackage{xcolor}         
\usepackage{graphicx}
\usepackage{subcaption}
\usepackage{amsmath}
\usepackage{todonotes}
\usepackage{wrapfig}

\title{Vision-Language Models as a Source of Rewards}

\author{%
    \thanks{Authors listed in alphabetical order. Contributions in \autoref{sec:contributions}. \newline Corresponding emails: \href{mailto:vmnih@google.com,feryal@google.com,harrischan@google.com}{\texttt{\{vmnih,feryal,harrischan\}@google.com}}}\quad  
    Kate Baumli \And
    Satinder Baveja \And
    Feryal Behbahani \And
    Harris Chan \And
    Gheorghe Comanici \And
    Sebastian Flennerhag \And
    Maxime Gazeau \And
    Kristian Holsheimer \And
    Dan Horgan \And
    Michael Laskin \And
    Clare Lyle \And
    Hussain Masoom \And
    Kay McKinney \And
    Volodymyr Mnih \And
    Alexander Neitz \And
    Dmitry Nikulin \And
    Fabio Pardo \And
    Jack Parker-Holder \And
    John Quan \And
    Tim Rocktäschel \And
    Himanshu Sahni \And
    Tom Schaul \And
    Yannick Schroecker \And
    Stephen Spencer \And
    Richie Steigerwald \And
    Luyu Wang \And
    Lei Zhang \AND
    {\normalfont{Google DeepMind}} \\
}

\begin{document}

\maketitle


\newcommand{\revisit}[1]{\textcolor{blue}{#1}}
\newcommand{\revise}[1]{\textcolor{blue}{#1}}
\newcommand{\review}[1]{\textcolor{blue}{#1}}
\newcommand{\alt}[1]{\textcolor{brown}{#1}}
\newcommand{\altalt}[1]{\textcolor{cyan}{#1}}
\newcommand{\althide}[1]{}
\newcommand{\added}[1]{\textcolor{red}{#1}} 
\newcommand{\addedok}[1]{\textcolor{cyan}{#1}}
\newcommand{\remove}[1]{\textcolor{green}{#1}}
\newcommand{\removemaybe}[1]{\textcolor{orange}{#1}}
\newcommand{\removehide}[1]{}



\newif\ifcomments
\commentstrue

\ifcomments
\newcommand{\HC}[1]{\textcolor{cyan}{({\bf HC:} #1)}}
\newcommand{\HCchide}[1]{}

\newcommand{\VM}[1]{\textcolor{purple}{({\bf VM:} #1)}}
\newcommand{\VMhide}[1]{}

\newcommand{\ML}[1]{\textcolor{magenta}{({\bf ML:} #1)}}
\newcommand{\MLhide}[1]{}

\else

\newcommand{\HC}[1]{}
\newcommand{\HChide}[1]{}

\newcommand{\VM}[1]{}
\newcommand{\VMhide}[1]{}

\newcommand{\ML}[1]{}
\newcommand{\MLhide}[1]{}

\fi

\begin{abstract}
Building generalist agents that can accomplish many goals in rich open-ended environments is one of the research frontiers for reinforcement learning. A key limiting factor for building generalist agents with RL has been the need for a large number of reward functions for achieving different goals. We investigate the feasibility of using off-the-shelf vision-language models, or VLMs, as sources of rewards for reinforcement learning agents. We show how rewards for visual achievement of a variety of language goals can be derived from the CLIP family of models, and used to train RL agents that can achieve a variety of language goals. We showcase this approach in two distinct visual domains and present a scaling trend showing how larger VLMs lead to more accurate rewards for visual goal achievement, which in turn produces more capable RL agents.
\end{abstract}

\section{Introduction}
Many of the biggest successes of reinforcement learning (RL, \citep{Sutton1998}) have been in domains where a clear reward function was readily available. Reward functions that have been successfully used include game win/loss \cite{Tesauro95,SilverHuangEtAl16nature,alphastarblog}, change in game score \cite{bellemare2012ale,mnih2013atari}, change in underlying state like forward motion \cite{lillicrap2019continuous} and negative distance to a desired state configuration. With these successful applications of RL to challenging tasks there has been growing interest in building generalist agents capable of achieving many challenging goals in rich environments.

One of the main limiting factors in applying RL to building generalist agents is the need for many reward functions for different goals. Building a reward function which, when maximized, will lead to the achievement of a particular goal can be challenging, time consuming, and hard to scale \citep{leike2018scalable,popov2017data}. This is true both in simulated environments, where determining whether an agent successfully achieved an abstract goal like building a house from the underlying state variables is difficult to express in code, and in the real world, where a reward has to be computed from observations. 
These challenges have put the spotlight on automatic ways of generating reward functions for training generalist agents \citep{leike2018scalable,yang2023foundation}.

One particularly promising direction that has emerged recently is the use of vision-language models (VLMs) for building reward functions in visual environments \citep{fan2022minedojo,du2023vision,mahmoudieh2022zsrm, cui2022zest}. VLMs trained on large datasets of paired images and text have been show to perform well on a variety of visual detection, classification and question answering tasks out of the box and can act as good initializations for finetuning on specialized tasks or datasets. Recent work showed that a pretrained CLIP model \citep{radford2021clip} finetuned on paired Minecraft videos and text can be used as an effective shaping reward for training agents to achieve hunting and foraging tasks in Minecraft \citep{fan2022minedojo}.
Similarly, \citep{du2023vision} showed how a pretrained Flamingo \citep{alayrac2022flamingo} VLM can be finetuned on visual question-answering data to produce an accurate visual success detector for a variety of natural language goals.

Encouraged by the success of these approaches we explore the feasibility of using off-the-shelf VLMs to derive accurate reward functions for language goals in visual environments. We propose a way of deriving a sparse binary reward for visual achievement of language goals from pretrained CLIP image and language embeddings and show how it can be used to train agents to achieve a variety of language goals in the Playhouse \citep{playhouse} and AndroidEnv \citep{toyama2021androidenv} visual environments. 
We believe that our work presents a compelling demonstration of how off-the-shelf VLMs can be used to train grounded language agents without the need for finetuning on environment specific data.

\section{Related Work}

\subsection{VLMs  Rewards}

A number of prior works have investigated using pretrained models such as CLIP as a reward function for RL. Most closely related to our work is MineDojo~\citep{fan2022minedojo} which first finetunes CLIP with Minecraft videos to form MineCLIP. Then MineCLIP is used as a dense shaping reward function in addition to a ground truth binary reward function for programmatic tasks where ground truth is available, except for creative tasks where no ground truth is available. 
The main differences between this work and MineDojo are that
(i) we use an off-the-shelf CLIP model,
(ii) we do not utilize any ground truth information during RL training.
and (iii) we train a single RL agent that can solve thousands of language-based goals. Similar to MineDojo, CLIP4MC \citep{ding2023clip4mc} finetunes CLIP on a curated dataset of Minecraft Youtube videos. CLIP4MC then uses PPO as the RL training backbone and the policy is trained on the CLIP4MC reward in the same form as MineCLIP reward together with the sparse completion where available. 
CLIP has also been used for reward-shaping simulated robotics domain \citep{cui2022zest,mahmoudieh2022zsrm,sontakke2023roboclip}. 
RoboCLIP \citep{sontakke2023roboclip} computes a sparse trajectory level reward as the cosine similarity score between the embedding of the trajectory video and the embedding of the text description of the task. 
VLM-RMs \citep{rocamonde2023vision} proposes goal-baseline regularization by projecting out irrelevant CLIP embedding space dimensions from the goal embedding using a secondary baseline text prompt embedding, applied to training a controller for a simulated humanoid robot. 
CLIP encoder was used for semantic intrinsic reward signal for curiosity-driven exploration in \citep{tam2022semantic}. 
VIPER \citep{escontrela2023video} uses the conditional log-likelihood of a frozen action-free video prediction model, such as VideoGPT \citep{yan2021videogpt}, summed with an entropy bonus as the reward signal for reinforcement learning agents without groundtruth task rewards in DeepMind Control Suite \citep{tunyasuvunakool2020dm_control}, Atari \citep{bellemare2012ale}, and RLBench \citep{james2020rlbench} tasks.

An orthogonal approach for scaling up reward modeling is to use Large Language Models (LLMs) to translate task specification into reward functions executable programs, resulting in sparse rewards \citep{yu2023languagetorewards} or dense rewards \citep{xie2023text2reward, ma2023eureka}. Earlier works also use LLMs itself as binary reward in negotiation games \citep{kwon2023reward}, collaborative human-AI interaction games \citep{hu2023language}, or dense reward based on the cosine similarity of a caption (hard-coded or learned) of transition and goals in a 2D Crafter environment \citep{du2023guiding, hafner2021benchmarking}.

\subsection{VLMs for Hindsight Relabeling}

Several works used VLMs for hindsight relabeling to do behaviour cloning. \citep{sumers2023distilling} uses Flamingo for hindsight relabeling and then behaviour cloning in the Playhouse environment (similar to our \emph{Lift} task). DIAL \citep{xiao2022robotic} uses CLIP finetuned on robotic demonstration and autonomous data that has been hindsight labeled by human annotators to further relabel a larger set of offline robotic trajectories, then also perform behaviour cloning a language-conditioned policy. 

\subsection{VLMs as Success Detectors}

SuccessVQA \citep{du2023vision} finetunes smaller 3B Flamingo models to act as success detectors by formulating the problem as a Visual Question Answering (VQA). Most of their experiments focus on how accurate the success detector is on offline datasets across various domains (robotics, Playhouse, Ego4D), with one experiment using the SuccessVQA model for reward-filtered behaviour-cloning.  

\begin{figure}[t]
    \centering
    \includegraphics[width=0.9\textwidth]{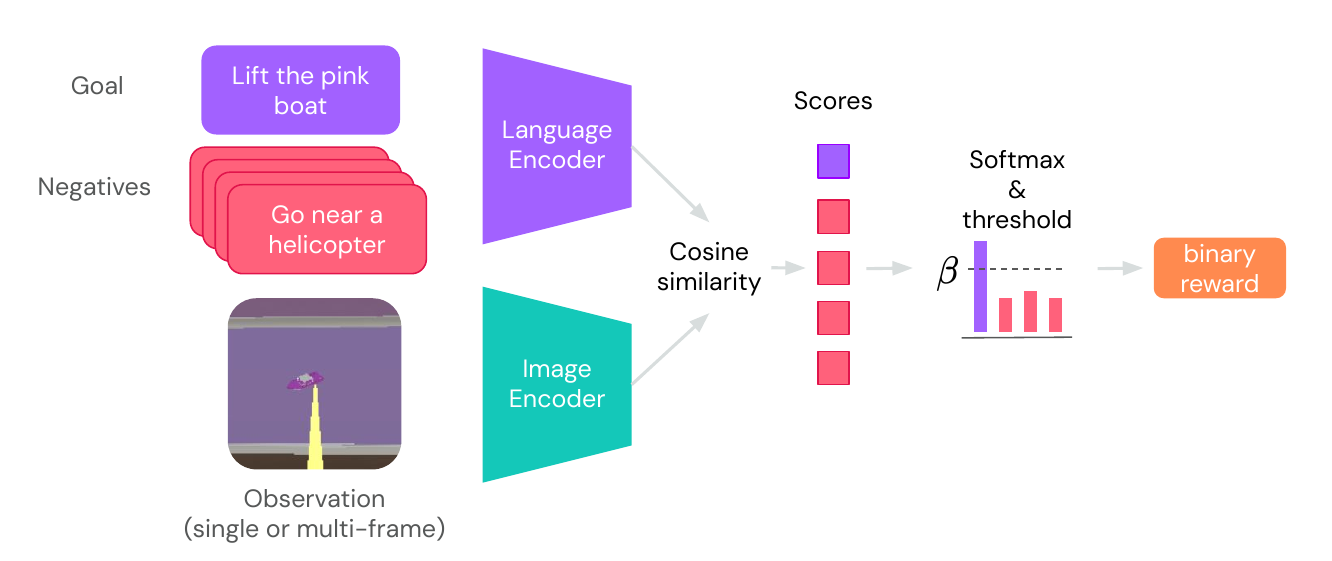}
    \caption{Architecture for Vision-Language Models (VLMs) as rewards. The VLM trained contrastively contains an image encoder $f_\theta$ and language encoder $g_\theta$. We embed the current environment observation frame(s) using the image encoder, along with the desired goal language descriptions $l$ and negative language descriptions using the language encoder. The reward is computed by taking the cosine similarity scores and applying softmax and thresholding. }
    \label{fig:vlm_reward_arch}
\end{figure}

\section{Method}

In this work we investigate how to leverage contrastive VLMs like CLIP to produce text-based reward models. Our aim is to produce a reward function that, given a language-based goal and an image observation, produces a scalar reward corresponding to whether or not that goal has been achieved. Concretely, we construct a reward model which consists of an image encoder $f_\theta (o)$ and text encoder $g_\theta(l)$ where $o \in \mathcal O$ is an observation and $l \in \mathcal L$ is a text-based goal. The reward model inputs both observation and goal $r(o, l)$ and outputs a binary reward: $1$ if the goal has been achieved and $0$ otherwise. 

We operate in a partial observed Markov Decision Process (POMDP) setting where the agent sees observations $o_t \in \mathcal O$, acts in an environment with actions $a_t \in \mathcal A$, observes the next observation according to the environment dynamics $o_{t+1} \sim \mathcal P (o_{t+1}| a_t, o_t)$, and receives
a reward $r_t \sim \mathcal R$ with a discount factor $\gamma \in (0,1]$. In addition to the standard POMDP, our agent is given a language-based goal $l \sim \mathcal{L}$ for the duration of the episode and the reward is computed via a VLM as a function $r_t = r(o_{t+1}, l)$. The episode terminates either on timeout $T$ or when $r_t = 1$ meaning that the goal has been achieved according to the intrinsic VLM reward function. Note that we assume no access to the ground truth reward during training, and only use the ground truth reward for evaluation.

\subsection{Turning CLIP into a reward function}
We first compute the probability $p_{o_t, l}$ that the agent achieves the goal given by language description $l$ after acting in the environment, out of a set of potential goals $l' \in \mathcal{L}$ in the task set $\mathcal{L}$, by applying softmax with temperature $\tau$ over the cosine similarity between the embedding of the state,  $f_\theta(o_{t+1})$, and the embedding of the language description $g_\theta(l)$ across the set of potential goals: 
\begin{align}
    p_{o_t, l} = \frac{\exp(f_\theta(o_{t}) \cdot g_\theta(l)/\tau)}{\sum_{l'} \exp(f_\theta(o_{t}) \cdot g_\theta(l')/\tau)}
\end{align}
We then convert this reward function into a binary reward function by thresholding the probability on the hyperparameter value $\beta \in [0,1]$:
\begin{align}
    r_t \equiv r(o_{t+1}, l) = \mathbb{I} [p_{o_{t+1}, l} > \beta]
\end{align}
where $\mathbb{I}$ is an indicator function. To avoid stochastic rewards, we sample negatives uniformly from the task set $l \in \mathcal L$ and fix them throughout the duration of RL training. 
In our experiments, the task set is a discrete set of language goals that are known to be achievable in the environment. We leave investigations on generalizing the negatives sampling from generative distribution, such as samples from an LLM, for future work. 

\section{Experiments}
In our experiments, we aim to answer the following questions:
\begin{enumerate}
    \item Does maximizing the VLM reward maximize the ground truth reward? (Section \ref{sec:exp_max_reward})
    \item How does scaling the size of the VLM affect the performance of the VLM reward? (Section \ref{sec:exp_scaling})
    \item Can online agent performance be predicted from offline evaluation metrics? (Section \ref{sec:exp_scaling})
    \item What are the effects of prompt engineering on the performance of the VLM reward? (Section \ref{sec:exp_prompt_eng})
\end{enumerate}

\subsection{Experimental Setup}

Our experimental setup is similar to that of standard online RL where an agent maximizes returns via trial-and-error interactions with an environment. The only difference is that rather than training against a ground truth or hand-engineered reward function, our agent trains against rewards emitted by a VLM. For evaluation purposes, we plot both the VLM rewards, which we call intrinsic, and ground truth rewards, which are provided by access to the simulator. We train a language-conditioned policy that is initialized randomly, while the reward model is a pre-trained CLIP VLM whose weights are frozen. We use Muesli~\citep{muesli} as the RL algorithm to optimize the policy against the VLM reward function. The environment samples a goal from the set of training goals, and the agent is tested on held-out goals during the evaluation. 

\subsection{CLIP architecture}

We use pre-trained CLIP \citep{radford2021clip} models for all experiments. We investigate two types of encoder architectures for CLIP. The first utilizes Normalizer-Free Networks (NFNets)~\citep{brock2021nfnet}, which uses a ResNet image encoder with adaptive gradient clipping in lieu of batch normalization. The second is Swin~\citep{liu2021swin, xie2021swincontrastive}, which is a hierarchical vision transformer architecture for image encoding. We use the BERT architecture~\citep{devlin2018} as the language encoder in all CLIP models. 

\begin{figure}[t]
    \centering
    \includegraphics[width=\textwidth]{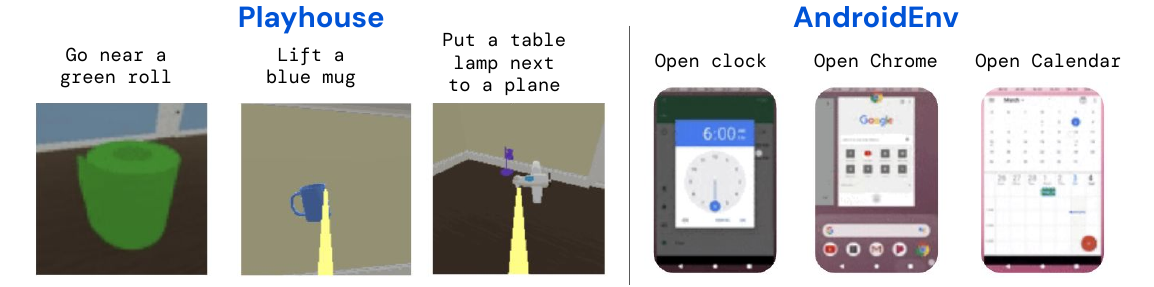}
    \caption{Environments and example tasks. (\textbf{Left}) Playhouse \citep{playhouse} consists of \textit{Find}, \textit{Lift}, and \textit{Pick and Place} tasks. (\textbf{Right}) AndroidEnv \citep{toyama2021androidenv} consists of opening app tasks across various apps on Android.}
    \label{fig:environments}
\end{figure}

\subsection{Environments and tasks}
We experimented with two domains for visual environments: (1) Playhouse \citep{playhouse}, and (2) AndroidEnv \citep{toyama2021androidenv}. Figure \ref{fig:environments} illustrates example observations from the two domains and their tasks. Playhouse is a Unity-based environment with an egocentric view inside a procedurally-generated home. We experimented with three families of tasks of increasing difficulty: (1) \textit{Find}, (2) \textit{Lift}, and (3) \textit{Pick and Place}. In the \textit{Find} task, the goal is to locate the object and approach the object as closely as possible. In the \textit{Lift} task, the agent must also grab the object and lift it. Finally, in the \textit{Pick and Place} task, the agent needs to pick up the desired object and place it near another specified object. 

AndroidEnv \citep{toyama2021androidenv} is an open-source environment built on the Android operating system, allowing the agent to interact through touchscreen gestures on an emulated Android device in real-time. is an Android device emulator that can be computer controlled. The observation space consists of realtime RGB images of the Android screen. The action space consists of swipe, touch, and type actions and mimics the way a human would interact with the device. We run experiments within two task sets within this environment. The first task, {\it Open Common Apps}, involves opening one of ten common apps such as Gmail, Google Sheets, Chrome, Calculator, Clock, Messages, Phone, Google Calendar, and Google Docs. The second task, {\it Open Diverse Apps}, expands the common app task set with fifty additional less well-known apps.

\begin{figure}[t]    
    \begin{subfigure}{0.33\textwidth}
        \includegraphics[width=\textwidth]{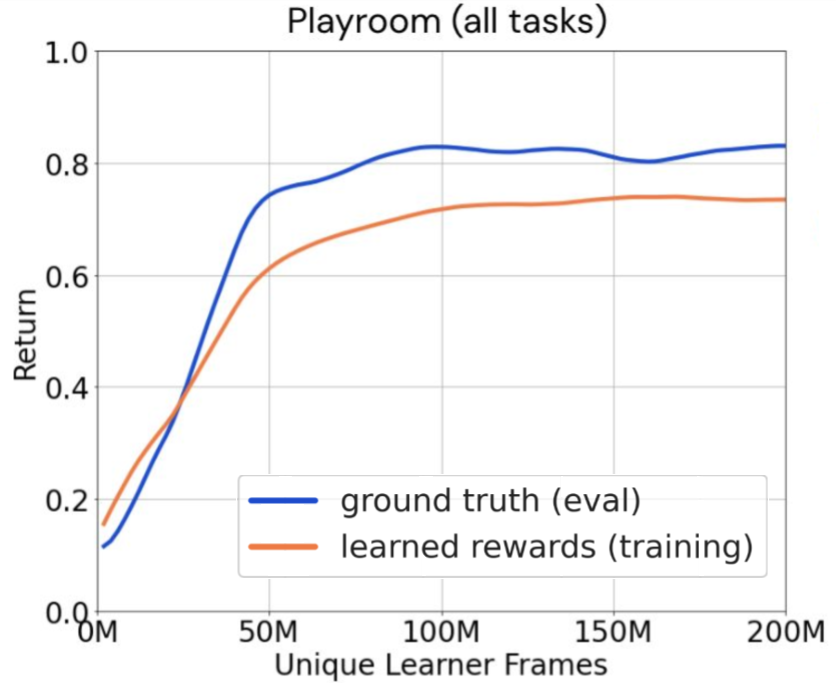}
        \caption{}
    \end{subfigure}
    \hfill
    \begin{subfigure}{0.315\textwidth}
        \includegraphics[width=\textwidth]{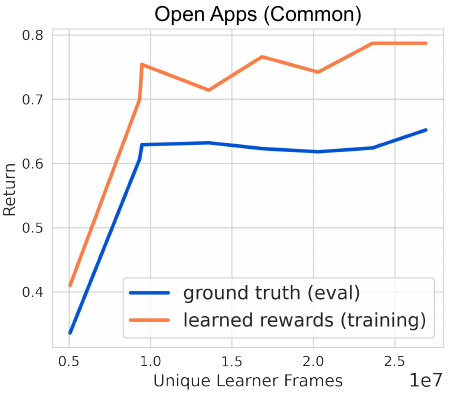}
        \caption{}
    \end{subfigure}
    \hfill
    \begin{subfigure}{0.32\textwidth}
        \includegraphics[width=\textwidth]{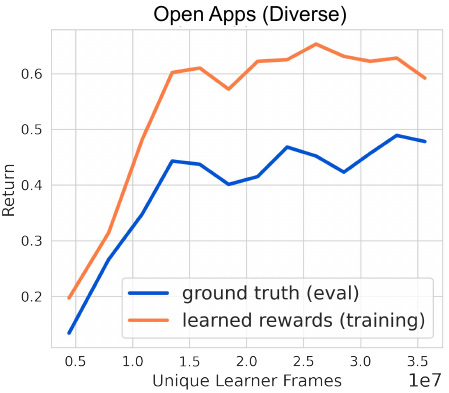}
        \caption{}
    \end{subfigure}
    \caption{Performance of an agent over the course of online reinforcement learning training when optimizing against the learned VLM reward. We measure both the (1) learned VLM reward return during training and (2) ground truth reward on held-out evaluation tasks. There is strong correlation between optimizing the learned VLM reward and the ground truth reward.}
    \label{fig:main_result}
\end{figure}

\begin{figure}[b]
    \hspace{-0.5cm}
    \begin{subfigure}{0.43\textwidth}
        \includegraphics[width=\textwidth]{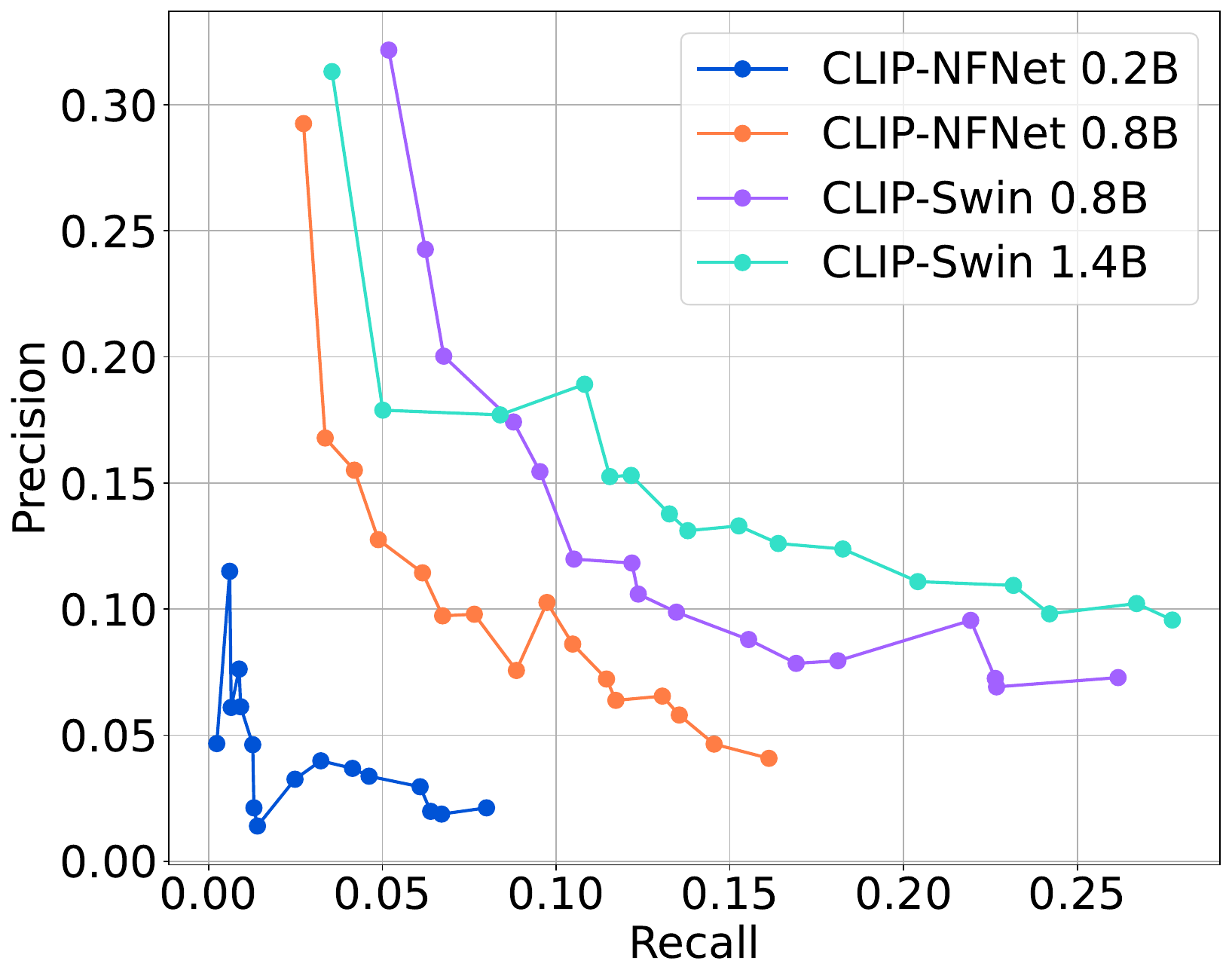}
        \label{fig:scaling_pr}
    \end{subfigure}
    \hspace{0.1cm}
    \hfill
    \begin{subfigure}{0.54\textwidth}
        \includegraphics[width=\textwidth]{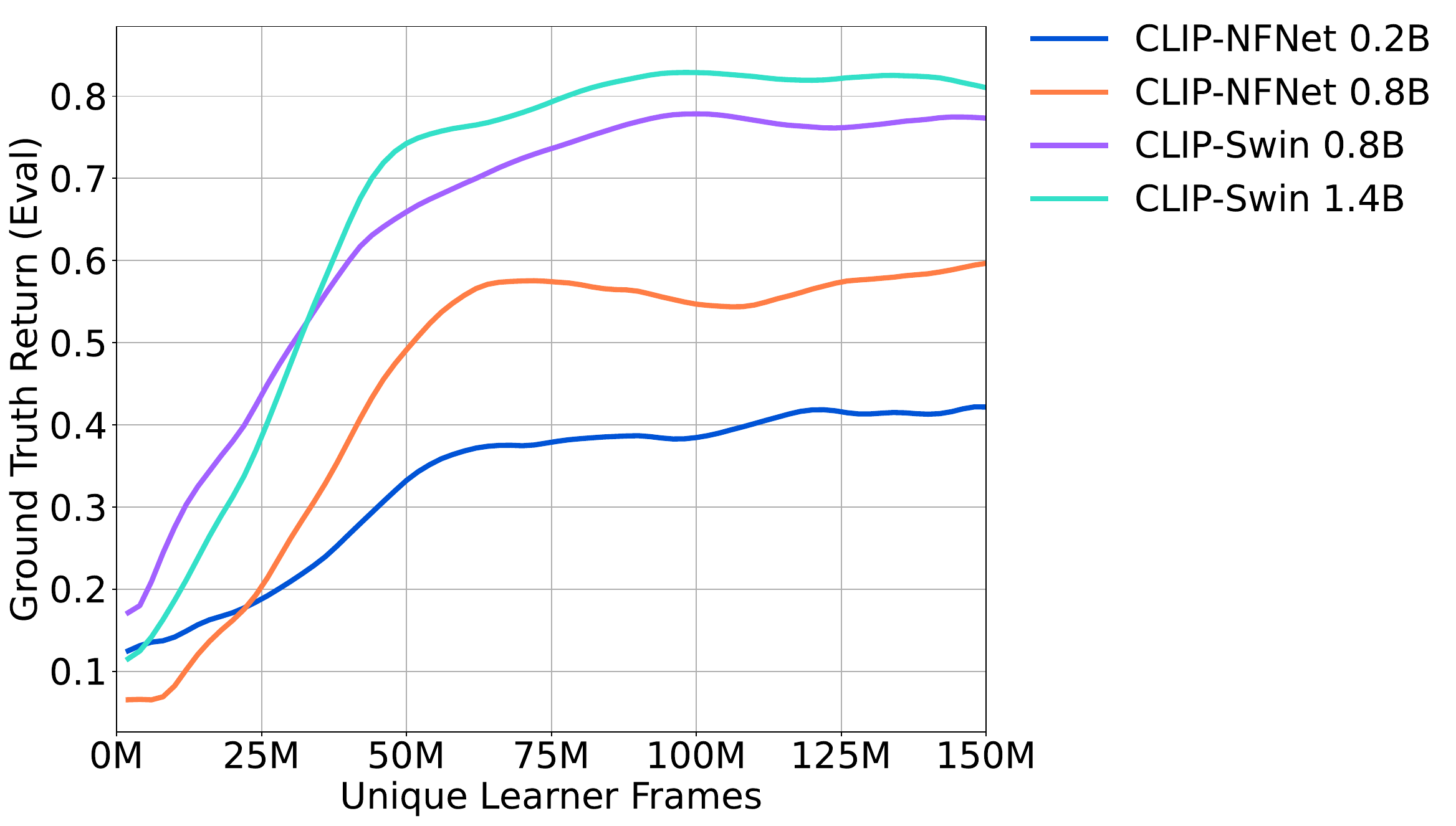}
        \label{fig:scaling_return}
    \end{subfigure}
    \caption{\textbf{Scaling reward model size}. (\textbf{Left}) Precision-Recall curves for varying VLM architecture and sizes on an offline fixed dataset of Playhouse trajectories. (\textbf{Right}) Ground truth returns on held-out evaluation tasks for Playhouse over the course of training with varying VLM reward sizes. }
    \label{fig:scaling}
\end{figure}

\subsection{Experimental Results}

\subsubsection{Maximizing learned rewards maximizes ground truth}
\label{sec:exp_max_reward}
For each of the environments and task families, we train a language goal-conditioned agent to maximize purely the VLM reward, without observing the ground truth reward. During the course of training, we also log the ground truth reward that the agent would have received. We observe in Figure \ref{fig:main_result} that the agent trained to maximize reward also maximizes the ground truth reward. Since the VLM reward is not perfectly accurate, there is a systematic gap between the VLM reward versus the ground truth reward. However, within the training budget, we did not observe reward hacking \citep{skalse2022defining} of our VLM reward, where the true reward drops off while the proxy VLM reward continues to increase.

\subsubsection{Scaling reward function model}
\label{sec:exp_scaling}
We investigate the scaling behaviour of the reward model as a function of the size of the underlying VLM. Firstly, we evaluate the accuracy of the VLM reward against ground truth binary reward on an offline, fixed dataset of Playhouse trajectories. Figure \ref{fig:scaling} (left) shows precision-recall curves obtained by a selection of models (CLIP with NFNet \citep{brock2021nfnet} and Swin \citep{liu2021swin,xie2021swincontrastive} vision encoders) evaluated against our Playhouse task set. The sensitivity is adjusted by varying the threshold $\beta$ above which the reward is given. We observe that increasing the size of the VLM used for the reward model (from 200M to 1.4B parameters) improves the precision-recall curves.
Figure \ref{fig:scaling} (right) shows the ground truth returns for held-out evaluation tasks over the course of training which are not given to the agent, when trained with VLM reward signals with different base models. We observe that the improved accuracy of the VLMs on offline datasets, when used as the only reward signal, does translate to better agent performance on ground truth evaluation metrics.

\subsubsection{Prompt Engineering VLM Reward}
\label{sec:exp_prompt_eng}

\begin{figure}[t]
    \begin{minipage}{0.35\textwidth}
        \begin{subfigure}{\textwidth}
            \includegraphics[width=\textwidth]{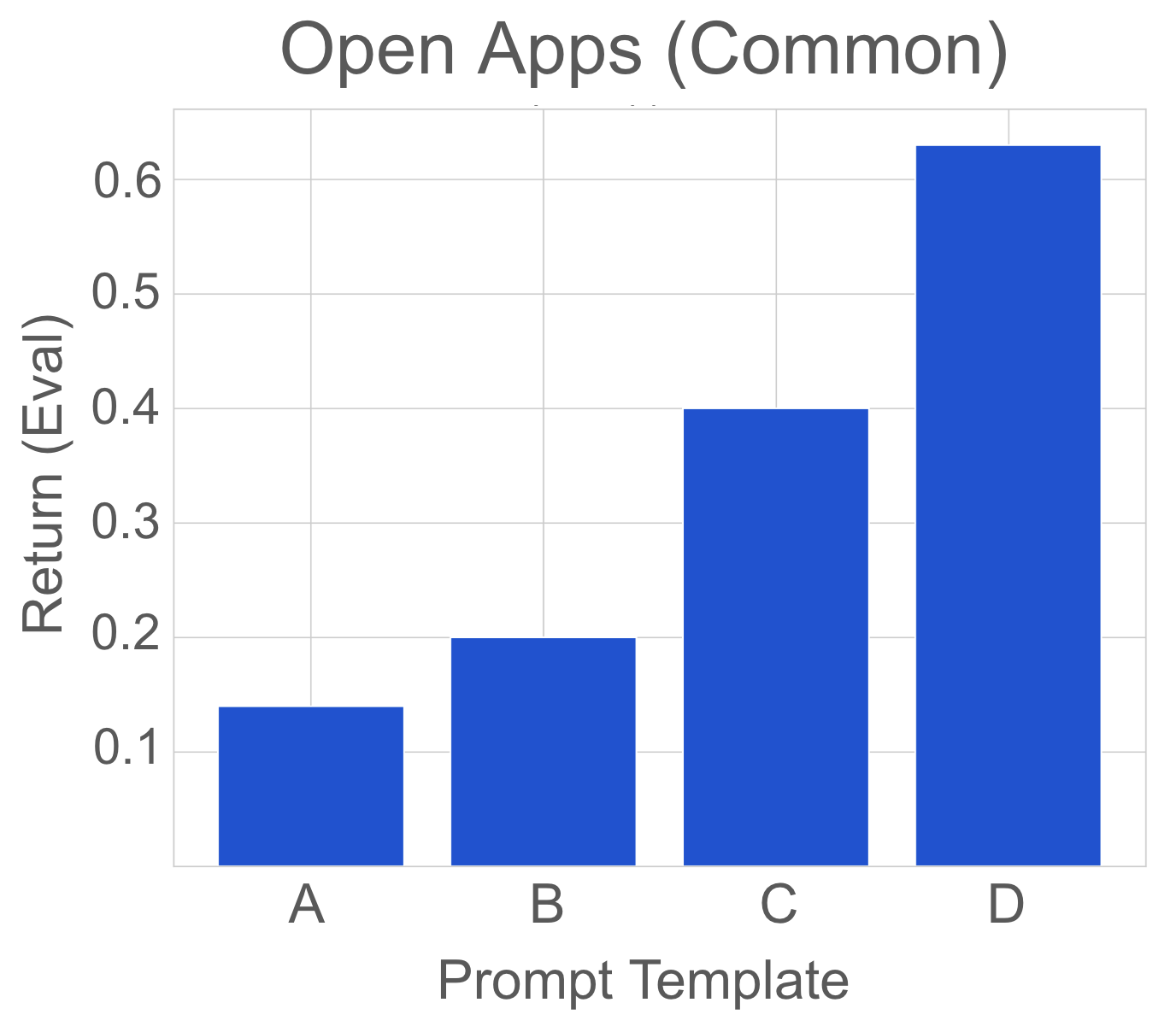}
            \label{fig:prompt_engineering_plot}
        \end{subfigure}
    \end{minipage}
    \hfill
    \begin{minipage}{0.54\textwidth}
        \centering
        \begin{subfigure}{\textwidth}
            \begin{tabular}{|c|c|}
                 \hline
                 & \textbf{Prompt Template} \\
                 \hline
                 A & \texttt{Open [TASK]} \\
                 \hline
                 B & \texttt{Open the [TASK] app} \\
                 \hline
                 C & \texttt{Screenshot of [TASK]} \\
                 \hline
                 D & \texttt{Screenshot of [TASK] on Android} \\
                 \hline
            \end{tabular}
            \label{fig:prompt_engineering_table}
        \end{subfigure}
    \end{minipage}
    \caption{(\textbf{Left}) Prompt engineering effects on the AndroidEnv \textit{Open App} task. More descriptive and specific prompts perform better when used as rewards. (\textbf{Right}) Prompt templates for the AndroidEnv \textit{Open App} tasks.}
    \label{fig:prompt_engineering}
\end{figure}

Manually prompt-engineering the text template has been found to improve the zero-shot classification of CLIP \citep{radford2021clip}. Similarly, we found that prompt-engineering the text template for our VLM reward can have a significant impact on the ground truth performance of the agent policies trained on the VLM reward. Figure \ref{fig:prompt_engineering}(left) compares the ground truth return evaluation of agent policies trained on various prompt templates (Figure \ref{fig:prompt_engineering} (right)). We hypothesize that, due to the training distribution of CLIP image captions, using action verbs such as ``\textit{open}'' in the prompt is not as helpful as providing a description of the successful states, such as ``Screenshot of [TASK] on Android''. 

\section{Conclusion}

This work shows how accurate rewards for visual achievement of language goals can be derived from off-the-shelf VLMs like CLIP without finetuning on domain-specific datasets. Our scaling trend analysis showed that the quality of the rewards derived from CLIP-style models increases as the number of parameters in the visual encoder grows. Taken together, we believe these results suggest that as the quality of VLMs improves it may be possible to train generalist RL agents in rich visual environments without the need for any task or domain-specific finetuning of the reward model.

\section{Authors and Contributions}
\label{sec:contributions}

We list authors alphabetically by last name. Please direct all correspondence to Volodymyr Mnih (\href{mailto:vmnih@google.com}{\mbox{vmnih@google.com}}), Feryal Behbahani (\href{mailto:feryal@google.com}{feryal@google.com}), and Harris Chan \\ (\href{mailto:harrischan@google.com}{harrischan@google.com}).

\subsection{Full-time Contributors:}

\begin{itemize}
\item \textbf{Kate Baumli:} agent research, infrastructure engineering
\item \textbf{Satinder Baveja:} advising
\item \textbf{Feryal Behbahani:} research vision, team leadership, agent research
\item \textbf{Harris Chan:} reward function research, paper writing
\item \textbf{Gheorghe Comanici:} agent research
\item \textbf{Sebastian Flennerhag:} agent research
\item \textbf{Maxime Gazeau:} reward function research
\item \textbf{Dan Horgan:} engineering leadership
\item \textbf{Michael Laskin:} reward function research, paper writing
\item \textbf{Volodymyr Mnih:} research vision, team leadership, reward function research, paper writing
\item \textbf{Alexander Neitz:} agent research, evaluation
\item \textbf{Fabio Pardo:} reward function research
\item \textbf{John Quan:} agent research
\item \textbf{Himanshu Sahni:} reward function research, evaluation
\item \textbf{Tom Schaul:} agent research
\item \textbf{Yannick Schroecker:} agent research, evaluation
\item \textbf{Stephen Spencer:} infrastructure engineering, evaluation
\item \textbf{Richie Steigerwald:} evaluation, infrastructure engineering
\item \textbf{Luyu Wang:} reward function research, infrastructure engineering
\item \textbf{Lei Zhang:} agent research
\end{itemize}

\subsection{Part-time Contributors:}

\begin{itemize}
\item \textbf{Kristian Holsheimer:} infrastructure engineering 
\item \textbf{Clare Lyle:} agent research
\item \textbf{Kay McKinney:} project management
\item \textbf{Hussain Masoom:} project management
\item \textbf{Dmitry Nikulin:} infrastructure engineering
\item \textbf{Jack Parker-Holder:} agent research
\item \textbf{Tim Rocktäschel:} advising

\end{itemize}
\bibliography{main}
\bibliographystyle{plainnat}

\appendix

\end{document}